\pdfoutput=1

\documentclass{article}
\usepackage{ijcai23}
\usepackage{times}
\usepackage{soul}
\usepackage{url}
\usepackage[hidelinks]{hyperref}
\usepackage[utf8]{inputenc}
\usepackage[small]{caption}
\usepackage{graphicx}
\usepackage{amsmath}
\usepackage{amsthm}
\usepackage{booktabs}
\usepackage[switch]{lineno}
\usepackage{booktabs}       
\usepackage{amsfonts}      
\usepackage{nicefrac}       
\usepackage{microtype}  
\usepackage{amssymb}
\usepackage{amsmath}
\usepackage{amsthm}
\usepackage{subcaption}
\usepackage{multirow}
\usepackage{pifont}
\usepackage{soul}
\usepackage{mathabx}
\usepackage{tikz}
\usepackage{todonotes}
\usepackage{cleveref}
\usepackage[ruled,vlined]{algorithm2e}
\usepackage{float}
\usepackage{placeins}
\usepackage{colortbl}
\usepackage{latexsym}

\theoremstyle{definition}
\definecolor{lightgray}{gray}{0.9}
\SetKwRepeat{Do}{do}{while}
\renewcommand{\vec}[1]{\mathbf{#1}}
\let\oldhat\hat
\renewcommand{\hat}[1]{\oldhat{\mathbf{#1}}}
\colorlet{verylightgray}{gray!9}
\colorlet{answerHighlightColor}{orange!30}
\renewcommand{\vec}[1]{\mathbf{#1}}

\newcommand{\F}{{\color{black}\ensuremath{\text{F}_1}}}
\newcommand*\circled[1]{\tikz[baseline=(char.base)]{
            \node[shape=circle,draw,inner sep=2pt] (char) {#1};}}
\title{Thought Flow Nets: From Single Predictions to Trains of Model Thought}

\author{
Hendrik Schuff$^{1,2}$
\and
Heike Adel$^1$
\And
Ngoc Thang Vu$^{2}$
\affiliations
$^1$Bosch Center for Artificial Intelligence, Renningen, Germany\\
$^2$Institut für Maschinelle Sprachverarbeitung, University of Stuttgart\\
\emails
\{Hendrik.Schuff,Heike.Adel\}@de.bosch.com,
Thang.Vu@ims.uni-stuttgart.de
}

\begin{document}

\maketitle

\begin{abstract}
When humans solve complex problems, they typically create a sequence of ideas (involving an intuitive decision, reflection, error correction, etc.) in order to reach a conclusive decision.
Contrary to this, today's models are mostly trained to map an input to one single and fixed output.
In this paper, we investigate how we can give models the opportunity of a second, third and $k$-th thought.
Taking inspiration from Hegel's dialectics, we propose the concept of a \emph{thought flow} which creates a sequence of predictions.
We present a \emph{self-correction mechanism} that is trained to estimate the model's correctness and performs iterative prediction updates based on the correctness prediction's gradient.
We introduce our method at the example of question answering and conduct extensive experiments that demonstrate (i) our method's ability to correct its own predictions and (ii) its potential to notably improve model performances.
In addition, we conduct a qualitative analysis of thought flow correction patterns and explore how thought flow predictions affect human users within a crowdsourcing study.
We find that (iii) thought flows enable improved user performance and are perceived as more natural, correct, and intelligent as single and/or top-3 predictions.
\end{abstract}

\section{Introduction}\label{sec:intro}
Today's classification models map a specific input $\vec{x}$, e.g., a token or a sentence, to an output $\hat{y}$ \cite{bishop_pattern_2006} where $\hat{y}$ can be, e.g., a class, a sequence (e.g., a generated text) or an answer span extracted from a context.
This mapping $\vec{x} \rightarrow \hat{y}$ might involve various modulations and abstractions of $\vec{x}$ in a latent space, e.g., hidden layers of a neural network, but typically does not allow variations or trajectories of $\hat{y}$.
Humans, on the other hand, rarely come to a single decision right-away but follow a complex thought process which involves reflecting on initial decisions, comparing different hypotheses or resolving contradictions.
While humans' trains of thought are extensively studied in cognitive sciences and philosophy---one particular example being Hegel's dialectics  \cite{maybee_hegels_2020}---such theories are rarely explored in machine learning.
However, with increasingly complex tasks that have large output spaces, such as question answering (QA)\footnote{A Longformer QA model can output 16M possible spans.}, or tasks that require multiple reasoning steps such as multi-hop QA, learning to directly hit the right prediction in one shot might be more difficult than to learn to iteratively self-correct an initial prediction.

\begin{figure}[t]
    \centering
    \includegraphics[width=0.9\columnwidth]{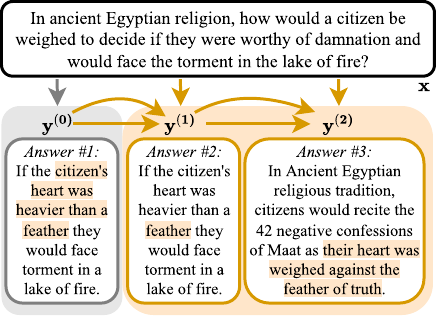}
    \caption{In contrast to the standard approach of mapping an input to an output in a single step (grey box), we propose a method that allows models to sequentially ``reconsider'' and update their predictions, i.e., the \textit{thought flow}. In this (real) question answering example, the orange box marks our thought flow extension, which corrects a flawed answer in two steps.}\label{fig:teaser}
\end{figure}

In this paper, we propose the concept of a \emph{thought flow} as a sequence of inter-dependent probability distributions.
Furthermore, we propose a simple \emph{correction module} to implement this concept.
It can be used on top of any model that provides output logits of one or multiple distributions.
In particular, it is inspired by the three moments of Hegel's dialectics which it relates to forward and backward passes of the model and is trained to judge whether the predicted class distribution corresponds to a correct prediction.

In our experiments on question answering, we
demonstrate our method's ability to self-correct flawed answer span predictions and identify qualitative patterns of self-correction, such as span reductions/extensions.
\Cref{fig:teaser} shows a real example of a thought flow that corrects a prediction ($\vec{y^{(0)}}$) that would be the output of a standard model to a new prediction ($\vec{y^{(2)}}$) within two steps, namely a shrinkage of the answer span and a cross-sentence answer jump.
We find that our method can achieve performance improvements up to 9.6\% \F (absolute) on a question answering dataset.

Finally, we assess the impact of thought-flow predictions on human users within a crowdsourcing study.
We find that thought-flow predictions are perceived as significantly more correct, understandable, helpful, natural, intelligent than single-answer predictions and/or top-3 predictions and result in the overall best user performance without increasing completion times or mental effort. 

To sum up, our contributions are (i) a formalization of a thought flow inspired from human thinking and Hegel's dialectics, (ii) a novel correction module and a corresponding gradient-based update scheme to generate a thought flow in a state-of-the-art transformer network, (iii) experiments on question answering that demonstrate its strong correction capabilities and identify qualitative patterns of self-correction, (iv) a crowdsourcing user study that demonstrates that thought flows can improve perceived system performance as well as actual user performance using the system.

\section{Thought Flow Networks}\label{sec:thought_flow_networks}
In this section, we present background on Hegel's dialectics (\Cref{subsec:background_hegel}), formalize thought flows based on it (\Cref{subsec:formalization}), and present a concrete implementation for question answering (\Cref{subsec:implementation}).

\subsection{Inspiration: Hegel's Dialectics}
\label{subsec:background_hegel}
To give models the opportunity to reflect and refine their predictions, we take inspiration from Hegel's dialectics.
Dialectics, in general, describes an argumentative method involving opposing sides \cite{maybee_hegels_2020}.
What distinguishes Hegel's dialectic from other dialectics is that in his dialectic, the opposing sides are views or definitions while, e.g., in Platon's dialectic the opposing sides are people \cite{maybee_hegels_2020}.
Besides its philosophical relevance, Hegel's dialectics has been related to various fields before, such as cognitive sciences \cite{riegel_dialectic_1973}, neuroscience \cite{boonstra_dialectics_2019} or optimization \cite{kadioglu_dialectic_2009}.

In the following, we will briefly introduce the three \textit{moments} of Hegel's dialectics and distinguish it from the thesis-antithesis-synthesis triad before we use them to derive our thought flow concept in the following section.

\paragraph{Three Moments.}
Hegel's dialectics distinguishes three moments:
(i) the \textit{moment of understanding}, (ii) the \textit{dialectical moment}, and (iii) the \textit{speculative moment}.
The moment of understanding refers to the initial, ``seemingly stable'' view.
In the second moment, this supposed stability is lost due to the view's one-sidedness or restrictedness and the initial determination \textit{sublates} itself into its own negation.
The speculative moment unifies the first two determinations by negating the contradiction \cite{maybee_hegels_2020}.

\paragraph{Thesis-Antithesis-Synthesis Triads.}
The three moments are often compared to a thesis-antithesis-synthesis triad, which was popularized by Heinrich Moritz Chalybäus, but \textit{cannot} necessarily be equated to it as argued by, e.g., Mueller~\shortcite{mueller_hegel_1958}.
While the thesis-antithesis-synthesis triad can suggest the notion of a ``one pass'' process, the dialectical process in Hegel's dialectic does not have to end after a single iteration, but can go through several iterations \cite{maybee_hegels_2020}.\footnote{A particular example of such an iterative process within Hegel's work can be found in the dialectical development of Hegel’s logic regarding the concepts of ``Abstract Purpose'' and ``Realized Purpose'' \cite{maybee_hegels_2020}.}
The possibility for iteration is an essential property of our thought flow 

\subsection{Formalization of Thought Flow Concept}
\label{subsec:formalization}
\begin{figure*}[t]
     \centering
     \begin{subfigure}[t]{.32\textwidth}
         \centering
     \includegraphics[width=\textwidth]{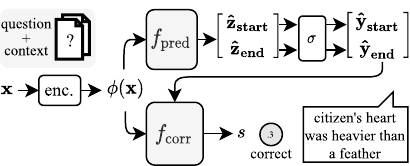}
     \caption{First label and correctness prediction ($\rightarrow$ moment of understanding).
     }\label{fig:step_1}
     \end{subfigure}%
     \hfill
     \begin{subfigure}[t]{.32\textwidth}
         \centering
     \includegraphics[width=\textwidth]{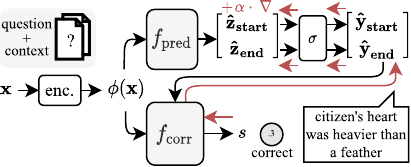}
     \caption{
     Gradient calculation w.r.t. the label logits ($\rightarrow$ dialectical moment).
     }\label{fig:step_2}
     \end{subfigure}%
     \hfill
     \begin{subfigure}[t]{.32\textwidth}
         \centering
     \includegraphics[width=\textwidth]{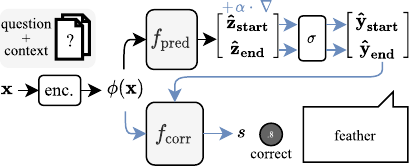}
     \caption{
     Update logits and correctness score ($\rightarrow$speculative moment).
     }\label{fig:step_3}
     \end{subfigure}%
     \caption{The steps of the prediction update scheme and their relation to the three moment of Hegel's Dialectics. The example shows the first answer change from \Cref{fig:teaser}.}\label{fig:steps}
 \end{figure*}

We now translate the abstract description of these three moments into a simplified mathematical setting that can be implemented in any (neural) model that uses a vector-valued representation of the input (such as an embedding) and outputs (tuples of) logits.
In particular, we embed Hegel's dialectics in a framework of obtaining an initial ``thought'' vector and iteratively updating it in the three ``moments''.
Note that our formalization is not to be understood as an accurate reflection of Hegel's dialectics but serves as a useful inspiration to enable the development of a novel machine learning model.

\paragraph{Thought.}
We model a \textit{thought} with $\vec{\hat{z}} \in Z$, the logits corresponding to a model's prediction and $Z \subseteq \mathbb{R}^c$ being the logit space.\footnote{We choose $\vec{\hat{z}}$ over $\vec{\hat{y}}$ because we can modify logits in energy space without having to normalize in probability space.}
This $\vec{\hat{z}}$ serves as a representation of the model's ``decision state'' as it captures information about the most probable output as well as possible alternatives and uncertainty.

\paragraph{Moment of Understanding.}
The first moment relates to an initial, seemingly stable view.
We model this with the initial value of $\vec{\hat{z}^{(0)}}$, obtained from applying the prediction function $f_{\text{pred}}: \Phi \rightarrow Z$ to the model to the encoded input $\phi(\vec{x})$ with an encoding function $\phi: \mathbb{R^d} \rightarrow \Phi$ and the encoding space $\Phi \subseteq \mathbb{R}^e$ (see \Cref{fig:step_1}).

\paragraph{Dialectical Moment.}
In the second moment, the stability breaks down due to the view's one-sidedness or restrictedness.
To model this, we first introduce a new function $f_{\text{corr}}: Z \times \Phi \rightarrow \mathbb{R}$ that differentiably maps $\vec{\hat{z}^{(0)}}$ to a correctness score $s \in \mathbb{R}$ that is an estimate of the quality of the model prediction corresponding to $\vec{\hat{z}^{(0)}}$ while being conditioned on $\phi(\vec{x})$.
Intuitively, $f_{\text{corr}}(\vec{\hat{z}^{(0)}}, \phi(\vec{x}))$ scores how good the current prediction corresponding to $\vec{\hat{z}^{(0)}}$ is given the model input corresponding to $\phi(\vec{x})$.
Next, we formalize the dialectical moment with the gradient of the correctness score with respect to $\vec{\hat{z}^{(0)}}$, i.e. $\nabla^T_{\vec{\hat{z}}^{(0)}} s$ (see \Cref{fig:step_2}).
Thus, we ask ``How does the thought $\vec{\hat{z}^{(0)}}$ have to change in order to be more correct?''
This gradient represents the view's instability:
As it creates a tension away from the current $\vec{\hat{z}^{(0)}}$ towards a new one, it destroys its stability and thus negates the initial view.

\paragraph{Speculative Moment.}
The third moment unites the initial view with the negation from the dialectical moment.
We formalize this by modifying $\vec{\hat{z}^{(0)}}$ with a step into the gradient's direction that yields 
\begin{equation}
    \vec{\hat{z}^{(1)}} := \vec{\hat{z}^{(0)}} + \alpha^{(0)} \cdot \nabla^T_{\vec{\hat{z}}^{(0)}} s \label{eq:gradient_update}
\end{equation}
where $\alpha^{(0)}$ is a, potentially dynamic, step width and $\vec{\hat{z}^{(1)}}$ again constitutes the subsequent first moment of the next iteration (see \Cref{fig:step_3}).

\paragraph{Iteration.}
Iterative application of the dialectical and the speculative moment yields a sequence of logits $\left(\vec{\hat{z}^{(k)}}\right)_{k=0}^{N}$ and predictions $\left(\vec{\hat{y}^{(k)}}\right)_{k=0}^{N}$.

In the following, we concretize this abstract formalization for the example of question answering.

\subsection{Implementation in Transformers for QA}\label{subsec:implementation}

\Cref{fig:steps} visualizes our formalization for the question answering example introduced in \Cref{fig:teaser}.
We now discuss QA-related implementation details.

\subsubsection{Choosing Parameters and Functions}
To apply our abstract thought flow method to a real model we have to (a) determine how we structure the model prediction logit vector $\vec{\hat{z}}$, (b) choose an input representation $\phi(\vec{x})$ (that is passed to $f_{\text{pred}}$ as well as $f_{\text{corr}}$), (c) choose a parametrization of the correctness score prediction function $f_{\text{corr}}$ and (d) define what the correctness score $s$ measures.
In the following, we describe how these aspects can be realized in a transformer-based QA model.

\paragraph{Composing $\vec{\hat{z}}$.}
In extractive QA, a typical approach to model answer span extraction from a context of $L$ tokens, is to use two probability distributions: (i) $\vec{\hat{y}}_{\text{start}} \in [0,1]^L$ that assigns a probability of being the start of the answer to each token in the context and (ii) a respective end token distribution $\vec{\hat{y}}_{\text{end}} \in [0,1]^L$.
To match our previously defined formalization, we define
$\vec{\hat{z}^{(i)}} := \begin{bmatrix}\vec{\hat{z}_{\text{start}}^{(i)}} & \vec{\hat{z}_{\text{end}}^{(i)}}\end{bmatrix}^{\text{T}}$    
which is linked to the respective probability distributions via the softmax function $\sigma$: 
\begin{equation*}
    \vec{\hat{y}^{(i)}} := \begin{bmatrix}\vec{\hat{y}_{\text{start}}^{(i)}} & \vec{\hat{y}_{\text{end}}^{(i)}}\end{bmatrix}^{\text{T}} = \begin{bmatrix}\sigma(\vec{\hat{z}_{\text{start}}^{(i)}}) & \sigma(\vec{\hat{z}_{\text{end}}^{(i)}})\end{bmatrix}^{\text{T}}.
\end{equation*}

\paragraph{Input Representation $\phi(\vec{x})$.}
In contrast to transformer-based
classification models that usually rely on the embedding of the [CLS] token, typical transformer-based QA models apply a linear function on top of each token's embedding that maps the embedding to a start and an end logit.
To represent the input, we thus need a choice of $\phi(\vec{x})$ that captures the relevant parts of the (potentially very long) input.
We choose a weighted average over all token embeddings.
As weights, we choose the element-wise product of the predicted start and end probabilities.
We thus define $\phi(\vec{x}) \in \mathbb{R}^d$ with $d$ denoting the dimension of the embeddings\footnote{E.g., 768 for BERT-base \cite{devlin_bert_2019}.} as:
\begin{align}
    \vec{\tilde w^{(i)}} := & \left( \vec{\hat{y}_{\text{start}}^{(i)}} \odot \vec{\hat{y}_{\text{end}}^{(i)}} + \epsilon \cdot \mathbf{1} \right) & \in \mathbb{R}^{L} \\
    \phi(\vec{x})^{(i)} := & \left[ e_1, e_2, ..., e_{L} \right] \cdot \frac{\vec{\tilde w^{(i)}}}{\Sigma_j \vec{\tilde w^{(i)}}_j} & \in \mathbb{R}^d
\end{align}
where $\epsilon$ is a small constant that ensures that we do not divide by zero, $e_i$ is the embedding of the $i$-th token and $\odot$ is element-wise multiplication.
The intuition behind this is that the correction module should have access to all information about the context that the prediction model focused on.

\paragraph{Choosing $f_{\text{corr}}$.}
We use a two-layer MLP with SELU activation \cite{klambauer_self-normalizing_2017} to map the concatenated vector
\begin{equation}
    \begin{bmatrix}\text{dropout}(\phi(\vec{x})^{(i)}) & \vec{\hat{z}^{(i)}_{\text{start}}} & \vec{\hat{z}^{(i)}_{\text{end}}}\end{bmatrix}^{\text{T}} \in \mathbb{R}^{d+2\cdot L} \label{eq:correction_input}
\end{equation}
to a correctness score $s$. 
Note that $f_{\text{corr}}$ does not receive the decoded answer text but directly uses the start and end logits to provide differentiability.

\paragraph{Correctness Score $s$.}
Following standard evaluation metrics for question answering, we use the \F-score of the predicted answer as the correctness score that $f_{\text{corr}}$ is trained to predict.

\subsubsection{Training}
To train $f_{\text{corr}}$, we freeze the parameters of $f_{\text{pred}}$. 
Then, we pass the training instances through the whole model (including $\phi$, $f_{\text{pred}}$ and $f_{\text{corr}}$) as shown in \Cref{fig:step_1} to obtain the predicted correctness score $s$ (i.e., $f_{\text{corr}}$ predicts an \F-score estimate without access to the ground-truth answer span).
We determine the ground-truth correctness score by calculating the \F-score
between the ground truth answer and the answer prediction from $f_{\text{pred}}$.
We define the correction prediction loss as the mean squared error between the calculated score and the predicted $s$ and train $f_{\text{corr}}$ to minimize it.

\subsubsection{Inference}
At inference time, we encode a new input and predict (i) the answer start and end logits using $f_{\text{pred}}$ and (ii) an estimated \F-score $s$ of the predicted answer span using the correction module $f_{\text{corr}}$ as shown in \Cref{fig:step_1}.
Instead of directly using the initial logits as the model's prediction --- as would be done in a standard model --- we iteratively update the logits w.r.t. the estimated correctness score's gradient following our formalization from \Cref{subsec:formalization} as shown in \Cref{fig:step_2,fig:step_3}.

\paragraph{Update Rule.}
As described in \Cref{subsec:formalization}, we aim at modifying $\vec{\hat{z}}^{(i)}$ such that the correction module assigns an increased correctness (i.e., \F-score in this application to QA).
To apply \Cref{eq:gradient_update}, we have to define how the step size $\alpha$ is chosen in our QA application.
We choose a time-independent $\alpha$ such that a predefined probability mass $\delta$ is expected to move.
To this end, we first take a probing step of length one, calculate the distance as the $L_1$ norm between the initial distribution and the probe distribution and choose the step width $\alpha \in \mathbb{R}^+$ such that it scales the linearized distance to  the hyperparameter $\delta$:
\begin{equation}
     \alpha := \left[\frac{\delta}{||\sigma(\vec{\hat{z}}^{(i)}) - \sigma(\vec{\hat{z}}^{(i)} + \nabla^T_{\vec{\hat{z}}^{(i)}} s)||_1 + \epsilon} \right] \label{eq:alpha}
\end{equation}
where $\sigma(\cdot)$ denotes the softmax function and $\epsilon \in \mathbb{R^+}$ is a small constant for numerical stability.

\paragraph{Monte Carlo Dropout Stabilization.}
The gradient $\nabla_{\vec{\hat{z}}^{(i)}} s$ is deterministic but can --- as we find in preliminariy experiments --- be sensitive to small changes in the input representation $\phi(\vec{x})$.
We therefore stabilize our correction gradient estimation by \textit{sampling} and averaging gradients instead.
For this, we use the dropped-out input encoding from \Cref{eq:correction_input} and sample five gradients for every step using MCDrop \cite{gal_dropout_2016}.

\section{Question Answering Experiments}
\subsection{Data, Model and Training}
\paragraph{Dataset.}
We choose the \textsc{HotpotQA} dataset (distractor setting) \cite{yang_hotpotqa_2018} to evaluate our models because it contains complex questions that require multi-hop reasoning over two Wikipedia articles.
In the distractor setting, the model is ``distracted'' by eight irrelevant articles that are passed to the model in addition to the two relevant articles.
In addition to yes/no/answer span annotations, \textsc{HotpotQA} also provides explanation annotations in the form of binary relevance labels over the paragraphs of the relevant articles which we do not use when training our models.
As the public test set is secret, we use the official validation set as test set and sample a custom validation set of size 10k from the training set leaving 80,564 training instances.

\paragraph{Base model.}
We use a Longformer-large \cite{beltagy_longformer_2020} model\footnote{\url{https://huggingface.co/allenai/longformer-large-4096}} with a linear layer on top that maps token embeddings to start and end logits as our underlying question answering model.
The model reaches 63.5\% \F~(SD=0.6) on the \textsc{HotpotQA} validation set averaged over three random seeds and can handle input lenghts up to 4096 tokens which enables us to feed-in the entire context as a single instance without truncation.
The model's input is a single token sequence that contains the question followed by the answer context (i.e., the 10 concatenated Wikipedia articles).
The model's output are two distributions over the input tokens (i.e., two 4096-dimensional distributions), one for the answer start position and one for the answer end position.
This allows the model to choose its answer from any text span within the context.
We prepend a ``yes'' and a ``no'' token to the context, that offers the advantage of modeling these answer options within the same distributions as the text span answers.
In total, this model has 435M parameters compared to the additional 331k parameters our MLP implementation of $f_{\text{corr}}$ adds. 

\begin{figure*}[t]
    \centering
    \begin{subfigure}[t]{.33\textwidth}
        \centering
    \includegraphics[width=0.9\textwidth]{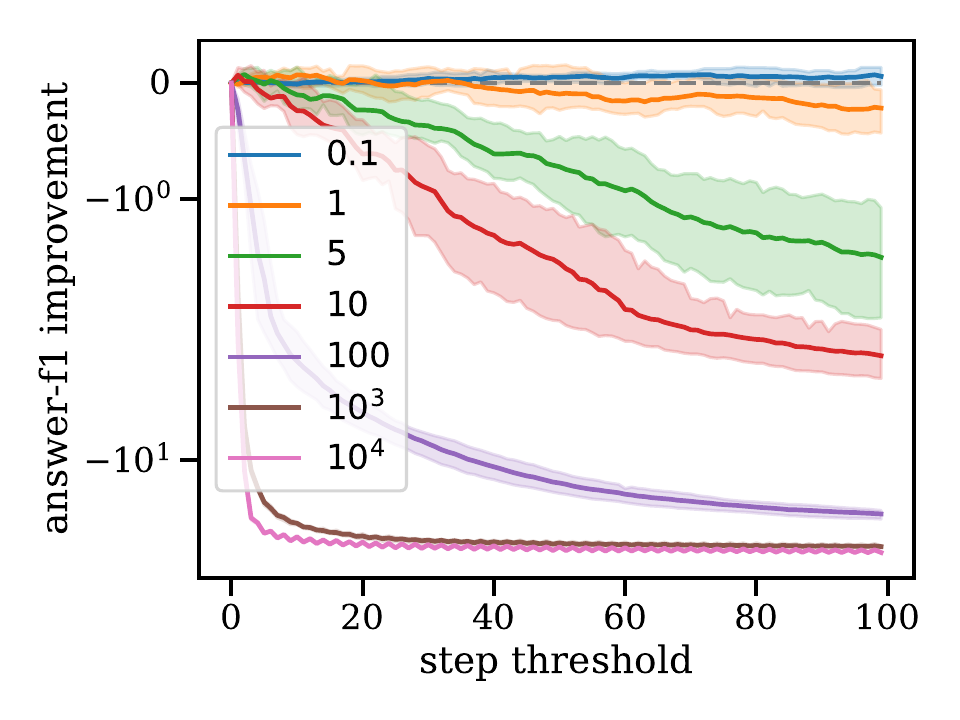}
    \caption{Non-oracle-stopped flows.}\label{fig:improvements_real}
    \end{subfigure}%
    \hfill
    \begin{subfigure}[t]{.33\textwidth}
        \centering
    \includegraphics[width=0.9\textwidth]{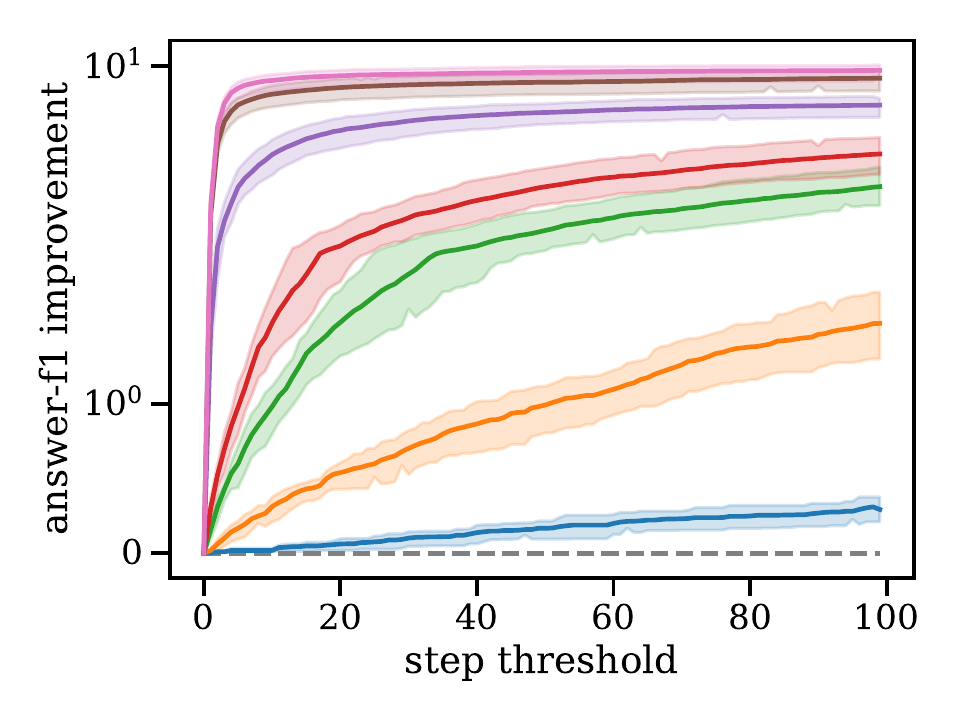}
    \caption{Oracle-stopped flows.}\label{fig:improvements_oracle}
    \end{subfigure}%
    \hfill
    \begin{subfigure}[t]{.33\textwidth}
        \centering
    \includegraphics[width=0.9\textwidth]{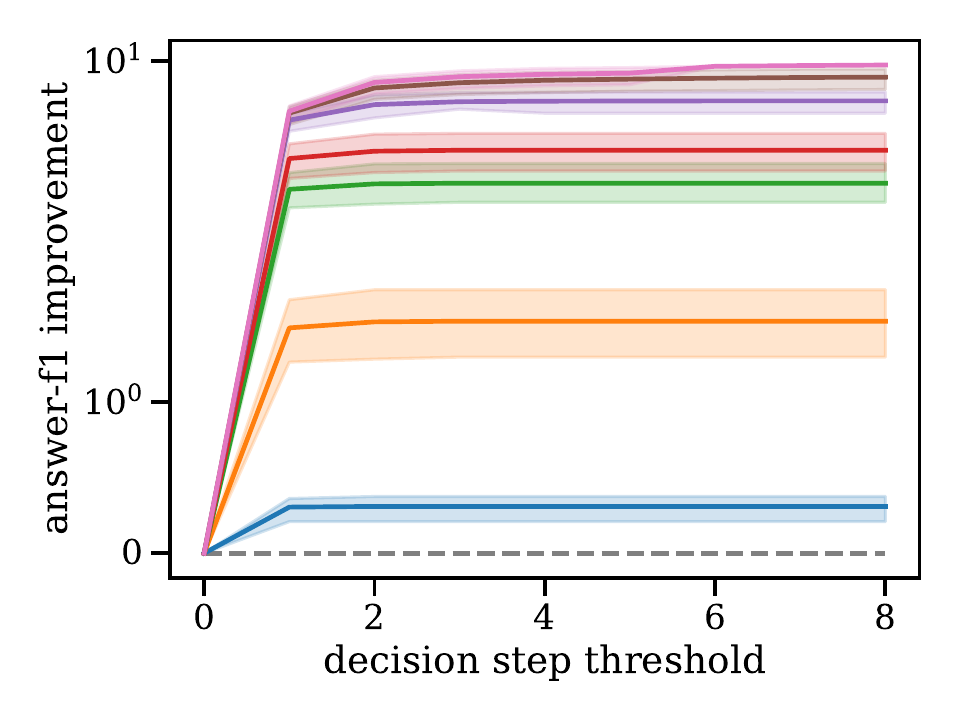}
    \caption{Oracle-stopped flows per decision change.}\label{fig:improvements_oracle_per_decision}
    \end{subfigure}
    \caption{Thought flows with different gradient scaling targets $\delta$ averaged over three seeds of a question answering model. Higher values for $\delta$ correspond to more aggressive decision changes. Without a stopping oracle that stops when the thought flow does no longer improve an answer (left), only $\delta=0.1$ provides consistently stable, but very small F1 improvements. With an oracle (middle), higher values for $\delta$ reach higher and faster F1 improvements up to $>$9\%. Nearly all performance gains are achieved by the first decision change (right). y axes use a symlog scale. Improvements are reported as absolute F1 scores (not relative to the the base performance).}\label{fig:improvements_real_and_oracle}
\end{figure*}

\paragraph{Training Details.}
We first train the base models on for five epochs on a single V100 GPU using a learning rate of $10^{-5}$, an effective batch size of 64 using an AdamW optimizer \cite{DBLP:conf/iclr/LoshchilovH19}, early stopping and a cross entropy loss on the start/end logits.
We subsequently train the correction modules using the same setting but the mean squared error loss function for \F-score prediction training.
Training one model each took approximately three days.
In the following, we report all results as averages over three random seeds including standard deviations.

\subsection{Performance Improvements}
\paragraph{How does Performance Vary Over Steps?}
\Cref{fig:improvements_real} shows how \F-scores per gradient scaling target $\delta$ evolve over 100 steps.
We observe that small $\delta$ values enable small \F~improvements.
While $\delta=0.1$ consistently improves \F-scores, all other $\delta$ values eventually deteriorate \F-scores.
The higher the $\delta$ value, the quicker the \F~decrease.
We conclude that (i) very small $\delta$ values are not sufficient to reach notable performance gains and that (ii) larger $\delta$ can initially improve performance but then ``overshoot'' with their corrections.
We hypothesize that a remedy to this trade-off is to use larger $\delta$ values but stop the corresponding flows at the right time.

\paragraph{What if We Had a Stopping Oracle?}
To test this hypothesis, we introduce an oracle stopping function that  stops the thought flow where it achieves it best \F~performance.
\Cref{fig:improvements_oracle} shows that, with this oracle function, thought flows can reach performance improvements up to 9.6\% \F (SD=0.61).

\Cref{fig:improvements_oracle_per_decision} shows that almost all performance improvements are due to the first decision change within the thought flows and answer spans constantly improve and do not randomly shift across the context.
This observation shows that single thought flow changes are highly effective and can reach substantial corrections fast.

\begin{table*}[t]
\centering
\resizebox{0.99\linewidth}{!}{
\begin{tabular}{p{0.06\linewidth}p{0.05\linewidth}p{1.1\linewidth}}
\toprule
\textbf{Pattern} & \textbf{Frequ.} & \textbf{Example} \\ \midrule
\quad\multirow{5}{*}{\centering\rotatebox[origin=c]{90}{\textit{cross-sentence}}} & \multirow{5}{*}{52.7\%} & Question: Who is older Danny Green or James Worthy?\\
& & (1) \colorbox{answerHighlightColor}{Daniel Richard "Danny" Green, Jr.} (born June 22, 1987) is an American professional basketball player for the San Antonio Spurs of the National Basketball Association (NBA).\\
& & (2) \textbf{\colorbox{answerHighlightColor}{James Ager Worthy}} (born February 27, 1961) is an American professional basketball coach and former player, commentator, television host, and analyst.\\
\midrule
\quad\multirow{5}{*}{\rotatebox[origin=c]{90}{\textit{span reduction}}} & \multirow{5}{*}{23.3\%} & Question: What philosophy related to creationism is Paul Nelson noted for? \\
& & (1) Paul A. Nelson (born 1958) is an American philosopher of science noted for his advocacy of \colorbox{answerHighlightColor}{young earth creationism and \textbf{intelligent design}}\\
& & (2) Paul A. Nelson (born 1958) is an American philosopher of science noted for his advocacy of young earth creationism and \textbf{\colorbox{answerHighlightColor}{intelligent design}}\\
\midrule
\quad\multirow{5}{*}{\rotatebox[origin=c]{90}{\textit{span extension}}} & \multirow{5}{*}{21.3\%} & Question: Ronald Reagan and George H. W. Bush both held which position in office? \\
& & (1) The presidency of Ronald Reagan began on January 20, 1981, when Ronald Reagan was inaugurated as \textbf{\colorbox{answerHighlightColor}{President} of the United States}, and ended on January 20, 1989.\\
& & (2) The presidency of Ronald Reagan began on January 20, 1981, when Ronald Reagan was inaugurated as \textbf{\colorbox{answerHighlightColor}{President of the United States}}, and ended on January 20, 1989.\\
\midrule
\quad\multirow{3}{*}{\rotatebox[origin=c]{90}{\textit{in-sent.}}} & \multirow{3}{*}{7.3\%} & Question: When was the stadium that held the 2015 Magyar Kupa demolished? \\
& & (1) The stadium was closed in \colorbox{answerHighlightColor}{2016} and demolished in \textbf{2017} to give place to the new Ferenc Puskas Stadium.\\
& & (2) The stadium was closed in 2016 and demolished in \textbf{\colorbox{answerHighlightColor}{2017}} to give place to the new Ferenc Puskas Stadium.\\
\midrule
\quad\multirow{4.5}{*}{\rotatebox[origin=c]{90}{\textit{logic hops}}} & \multirow{4.5}{*}{4\%} & Question: Is the Pakistan fast bowler who joined the Kent County Cricket Club in June, 2011 a left-hand or right-hand batsmans? \\
& & (1) Wahab Riaz (Punjabi, Urdu:  ; born 28 June 1985) is a \colorbox{answerHighlightColor}{Pakistani} cricketer.\\
& & (2) He is a \colorbox{answerHighlightColor}{left-arm fast bowler and a \textbf{right-hand}} batsman.\\
\bottomrule
\end{tabular}}
\caption{A subset of correction Patterns identified in 150 randomly sampled thought flows using $\delta=1$. The \textbf{correct answer} is marked bold, the  \colorbox{answerHighlightColor}{predicted answer per flow step} is marked in orange. We provide the full list of identified patterns in the appendix.}\label{tab:qa_flows_qualitative}
\end{table*}

\subsection{Thought Flow Patterns}
In a qualitative evaluation, we identify thought flow patterns.
We randomly sample 150 instances from the subset of the official validation split for which the thought flow changed the initial answer prediction.
We identify six (non-exclusive) correction patterns and show selected examples in \Cref{tab:qa_flows_qualitative}.

\paragraph{Cross-Sentence.} With 52.7\%, this is the most frequent type of correction. The thought flow shifts the predicted answer from one sentence to another.
\paragraph{Span Reduction.} The thought flow can shorten the predicted answer span to correct the answer.
\paragraph{Span Extension.} Similarly, the thought flow can also enlarge a predicted answer span to correct it.
\paragraph{In-Sentence.} On top of in-sentence span reduction/extension, the thought flow can also jump between non-overlapping spans within a sentence.
\paragraph{Entity Refinement.} In this correction pattern, the though flow keeps predicting the same entity but jumps to an alternative mention of the entity.
\paragraph{Logic Hops.} The thought flow performs a step-wise reasoning that first resolves the first step of HotpotQA's two-step reasoning structure before jumping to the second step, i.e., the correct answer.
\paragraph{Combinations.} We observe various combinations of the aforementioned patterns. A model can, for instance, jump between sentences, refine entities and reduce the answer span.

Corrections can also occur sequentially.
We provide detailed examples in the appendix.
We additionally observe flow patterns with very high number of decision changes.
These typically correspond to two- or three-cycles between answer spans or exhibit a seemingly chaotic behavior. 

\section{Human Evaluation}
While the previous section showed that thought flows can enable complex self-correction and can reach promising performance gains, we now investigate how thought flow predictions affect human users in an AI-assisted question answering task.

\subsection{Experiment Design}
We choose a within-subject design in which each participant is exposed to three variations of a question answering system.

\paragraph{Conditions.}
We aim at assessing the effect of the thought flow concept on users and therefore present the outputs of the oracle-stopped thought flow in one condition (\textbf{\textsc{TF}}) and compare it to two baseline conditions.
As baselines, we use top-1 predictions (\textbf{\textsc{single}}) (to compare against standard models) and top-3 predictions (\textbf{\textsc{top-3}}) (to compare to an alternative approach to show several predictions).
For all conditions, we present the predicted answer(s) along with the sentence in which they appear in the context.

\paragraph{Dependent Variables.}
We study the effect of the condition (\textsc{single}, \textsc{TF} and \textsc{top-3}) on a set of dependent variables.
We include variables on a per-question level (after each question) and on a per-system level (after all questions of one condition).
The per-question variables include: 
(i) \textbf{human answer correctness}, (ii) \textbf{perceived model correctness} (iii) \textbf{perceived understanding}, (iv) \textbf{perceived helpfulness} and (v) \textbf{completion time}.
The per-system variables include:
(vi) \textbf{usability} using the UMUX questionnaire \cite{finstad_usability_2010,finstad_response_2013},
(vii) \textbf{mental effort} using the Paas scale \cite{paas1992training},
(viii) \textbf{anthropomorphism} using the respective subscale of the Godspeed questionnaire \cite{DBLP:journals/ijsr/BartneckKCZ09}\footnote{We drop the robotics-specific item regarding ``moving rigidly/elegantly'' as it is not applicable to question answering.},
(ix) \textbf{perceived intelligence} using the subscale from the same questionaire,
(x) \textbf{average completion time}.
We provide a list of all questionnaires in the appendix.

\paragraph{Apparatus.}
We sample 100 instances from the \textsc{HotpotQA} validation instances for which a thought flow using $\delta=1$ causes at least one prediction change.\footnote{If there is no prediction change, \textsc{TF} is identical to \textsc{single}.}
From these, we sample 30 instances per participant and randomly assign the instances to three bins of 10 questions (one bin per condition).\footnote{We statistically account for random effects of single questions.
}
We balance the six possible condition orders across participants and include three attention checks per participant.
We provide screenshots of the study interface in the appendix.

\subsection{Quantitative Results}
We use MTurk to recruit US crowdworkers with $>$90\% approval rate and the MTurk Masters qualification an collect responses from 55 workers.\footnote{We filter out two participants that did not pass the attention checks and replace them with two additional responses.}

\subsubsection{Statistical Models.}
\paragraph{Per-System Ratings.}
We analyze the per-system ratings using Friedman tests to account for the paired responses due to the within-subject design.\footnote{Although aggregated Likert item scores are commonly considered interval responses, we use (non-parametric) Friedman tests that only require ordinal responses and are more conservative than their parametric counterparts RM-ANOVAs.}
We use Holm-corrected Conover post hoc tests to identify significant pairwise differences.

\paragraph{Per-Item Ratings.}
Note that the within-subject design of our study possibly introduces inter-dependencies within ratings that we have to account for using an appropriate statistical model.
Additionally, our dependent variables are measured on different levels, e.g., completion time is measured on a ratio scale while human answer correctness is measured on a nominal (dichotomous) scale.\footnote{We follow related work and treat Paas mental effort, UMUX and Godpseed subscale responses as interval data but analyze single-item perceived understanding and helpfulness on an ordinal level.}
We therefore use (generalized) linear mixed models (GLMM) and cumulative link mixed models (CLMM) to (i) account for random effects of question and subject IDs, and (ii) account for the variables' respective measurement scales.
\footnote{We use (G)LMMs to analyze continuous and dichotomous responses (Gamma/binomial link) and CLMMs to analyze ordinal ones.}
We use LRT tests between the full model and the model without the condition variable to identify main effects of the condition variable and conduct Holm-corrected Tukey post-hoc tests. 

\begin{table*}
\centering
\resizebox{0.875\textwidth}{!}{%
\begin{tabular}{lcccccccccc}
\toprule
Condition           & \multicolumn{7}{c}{perceived quality}                                                                                 & \multicolumn{2}{c}{user performance}  \\
                    \cmidrule(lr){2-8}                                                                                                      \cmidrule(lr){9-10} 
                    & \cellcolor{green!25}correct$^{*}$ & \cellcolor{cyan!25}understand$^{*}$  & \cellcolor{cyan!25}helpful$^{*}$ & \cellcolor{cyan!25}usability & \cellcolor{cyan!25}mental effort & \cellcolor{green!25}humanlike$^{*}$  & \cellcolor{cyan!25}intelligent$^{*}$ & \cellcolor{cyan!25}time$^{*}$    & \cellcolor{green!25}answer F1$^{*}$       \\
\textsc{single}     & A             & A             & A             & \circled{A}   & \circled{A}   & A             & A             & \circled{A}   & A               \\
\textsc{top-3}      & A             & \circled{B}   & \circled{B}   & \circled{A}   & \circled{A}   & \circled{AB}  & \circled{B}   & B             & B               \\
\textsc{TF}  & \circled{B}   & \circled{B}   & \circled{B}   & \circled{A}   & \circled{A}   & \circled{B}   & \circled{B}   & \circled{AB}  & \circled{C}     \\
\bottomrule
\end{tabular}
}
\caption{Statistical results of our human evaluation ($N=55$).
``$^{*}$'' marks dependent variables on which a significant effect of the system condition was observed (Friedman tests and LRT tests for GLMM/CLMM).
Pairwise differences between conditions (Holm-adjusted Tukey/Conover tests) are reported as compact letter display codings.
E.g., the ``humanlike'' column shows that the post hoc test detected a significant difference between \textsc{single} and \textsc{TF} but no significant difference between any other pair. Similarly, the last column shows pairwise differences between all conditions and the \textsc{TF} condition reaches significantly higher human answer \F-scores than any other conditions.
Variables for which \textsc{TF} is among the best performing models are marked \colorbox{cyan!25}{cyan}, variables for which it is found to be the sole superior system are marked \colorbox{green!25}{green}.}\label{tab:human_eval_stats}
\end{table*}

\subsubsection{Results.}
We find significant differences for all dependent variables except usability and mental effort.
We summarize the results of our statistical analysis in \Cref{tab:human_eval_stats} using CLD codings \cite{piepho2004algorithm}.
We provide the detailed $p$ values for main efects and each pairwise comparison in the appendix.
In the following we discuss our findings for each dependent variable for which we found a significant main effect.

\paragraph{Perceived Answer Correctness.}
While there is no statistically significant difference between showing users single answers or top-3 predictions, displaying thought flows leads to significantly higher answer correctness ratings.
\paragraph{Understanding.}
Top-3 as well as thought flow predictions significantly increased the feeling understanding how the system came up with its answer compared to single predictions.
\paragraph{Helpfulness.}
Similarly, top-3 and the thought flow predictions significantly improve perceived system helpfulness compared to single predictions.
\paragraph{Anthropomorphism.}
While we observe no signficant difference in antropomorphism ratings between single and top-3 predictions, the thought flow predictions are perceived significantly more human-like/natural than the single answers.
\paragraph{Perceived intelligence.}
Both, top-3 and the thought flow predictions, lead to an significantly increased perceived system intelligence.
\paragraph{Completion Time.}
We observe that the top-3 predictions significantly improve completion times compared to single answers, but there is no significant increase for thought flows.
\paragraph{User Performance.}
While top-3 predictions already improve user performance in terms of \F-score of the user's given answer, thought flow predictions enable even higher performances, that are significantly higher than answers given in the single answer or top-3 conditions.
We additionally analyze user answers using exact match scores and find the same effects and model orders.

Overall, our results indicate that \textbf{thought flows are better or equally good than single answer or top-3 predictions regarding all evaluated dimensions.}
In particular for perceived answer correctness, humanlikeness and user performance, thought flows are significantly better than both, the single answers and the top-3 predictions.
While comparable (statistically indistinguishable) improvements of understanding, helpfulness, naturalness and intelligence can also be achieved using top-3 predictions, these come at the cost of significantly increased completion times compared to single answers.
In contrast, we do not find a significant time increase using thought flows.

\section{Related Work}\label{sec:related_work}

\paragraph{Cognitive Modeling and Systems.}
The fields of cognitive modeling and cognitive systems provide numerous models of human thinking \cite{rupert_cognitive_2009,busemeyer_cognitive_2010,levine_introduction_2018,lake_building_2017}.
While work in these fields often orients towards accurate descriptions of human cognition, our method does not aim to provide a plausible description of cognitive process but, instead, aims to apply a philosophical concept to machine learning in order to improve classification performance and user utility.

\paragraph{Confidence Estimation and Model Corrections.}
Estimating a model's confidence and the correctness of its predictions is addressed with various methods, including the training of secondary models for predicting the main model's uncertainty \cite{blatz_confidence_2004,devries_learning_2018}.
Among those, \textit{ConfidNet} is particularly related to our approach as it predicts the true-class probability of the main model \cite{corbiere_addressing_2019}.
In contrast, our correction module receives the class probabilities of the main model as an input and predicts a correctness score.
In difference to methods aiming at estimating accurate confidence scores, we predict such scores only as an auxiliary task in order to generate a gradient that allows us to update the model prediction.
Regarding model correction, the arguably most established approach to learn corrections of model predictions is gradient boosting \cite{friedman_greedy_2001} including its popular variant XGBoost \cite{chen_xgboost_2016}.
In contrast to those works, we do not use an ensemble of weak learners but propose a lightweight correction module that is applicable on top of any existing classification model.
Further, in our method, the correction module receives the main model's predictions and is able to directly adapt them.

\paragraph{Sequences of Predictions.}
The idea of iteratively predicting and correcting has been explored for a long time.
Early work includes Mori \textit{et al.} who present a non-neural iterative correction method tailored to estimate elevation maps from aerial stereo imagery  \cite{mori_iterative_1973}.
Katupitiya \textit{et al.} propose to iterate two neural networks to address the problem of predicting inputs of a mechanical process given the outputs of the process \cite{katupitiya_neural_2005}.
While their method is specifically designed for the task of input prediction, our work presents a general-purpose classification model that iterates class label predictions.
Besides those task-specific methods, there are models and inference methods that make use of an iterative prediction process by design, such as
Hopfield networks \cite{hopfield_neural_1982} and their modern variants \cite{barra_new_2018,ramsauer_hopfield_2020}, or Loopy Belief Propagation, Markov Chain Monte Carlo or Gibbs sampling \cite{bishop_pattern_2006,koller_probabilistic_2009}.
While these techniques can be linked to our work conceptually, they all require to train a new model.
In contrast, our approach can be applied to an existing neural model as well.
Another related approach is chain-of-thought prompting \cite{DBLP:journals/corr/abs-2201-11903} in which a language model is prompted with demonstrations of problem decomposition/reasoning in a few-shot manner and subsequently can be observed to show similar behavior in its answer.
While this method yields impressive model answers, it predicts \textit{one} answer that contains information on its deduction without changing or correcting its answer.
In contrast, our method is not targeted towards decomposition/reasoning but predicts a sequence of answers with the goal of iteratively improving it.

\paragraph{Learning to Stop.}
A further line of work, ACT \cite{DBLP:journals/corr/Graves16} and PonderNet \cite{DBLP:journals/corr/abs-2107-05407}, trains recurrent networks to learn when to stop applying recurrent transformations within the model.
While their approaches require the model to contain recurrent modules and to retrain the base model, our method only requires the model to yield output logits and leaves the base model unchanged.

\section{Conclusion}\label{sec:conclusion}
In this paper, we introduced  a task-agnostic self-correction formalism that turns a model's single output prediction into an evolving sequence of predictions---the \emph{thought flow}.
We take inspiration from Hegel's dialectics and propose a correction module along with a gradient-based update rule that sequentially updates a model's output distributions in the direction of an increasing self-estimate of correctness.
We apply our method to question answering models and conduct extensive experiments including human evaluation.
We find that thought flows (i) can increase \F-scores up to 9.3\%, (ii) exhibit complex self-correction patterns and (iii) provide significant improvements in human interaction and system perception including task performance and perceived system correctness and naturalness. A potential next step to further improve performance is learning to stop.

\appendix

\bibliographystyle{named}
\bibliography{references}

\clearpage
\onecolumn
\appendix
\section{Question Answering Experiments}\label{sec:qa_experiment_details}
\subsection{Dataset Details.}
We use the \textsc{HotpotQA} dataset \cite{yang_hotpotqa_2018}, which is an English multi-hop question answering data set.
It covers 90,564 training instances, 7,405 test validation instances and 7,405 test instances per setting (there are a distractor and a fullwiki setting).
Training instances are grouped by difficulty and cover 18,089 easy, 56,814 medium and 15,661 hard questions.
We refer to \cite{yang_hotpotqa_2018} for more details.

\begin{table*}[t]
\centering
\resizebox{0.99\linewidth}{!}{
\begin{tabular}{p{0.06\linewidth}p{0.05\linewidth}p{1.1\linewidth}}
\toprule
\textbf{Pattern} & \textbf{Frequ.} & \textbf{Example} \\ \midrule
\quad\multirow{5}{*}{\centering\rotatebox[origin=c]{90}{\textit{cross-sentence}}} & \multirow{5}{*}{52.7\%} & Question: Who is older Danny Green or James Worthy?\\
& & (1) \colorbox{answerHighlightColor}{Daniel Richard "Danny" Green, Jr.} (born June 22, 1987) is an American professional basketball player for the San Antonio Spurs of the National Basketball Association (NBA).\\
& & (2) \textbf{\colorbox{answerHighlightColor}{James Ager Worthy}} (born February 27, 1961) is an American professional basketball coach and former player, commentator, television host, and analyst.\\
\midrule
\quad\multirow{5}{*}{\rotatebox[origin=c]{90}{\textit{span reduction}}} & \multirow{5}{*}{23.3\%} & Question: What philosophy related to creationism is Paul Nelson noted for? \\
& & (1) Paul A. Nelson (born 1958) is an American philosopher of science noted for his advocacy of \colorbox{answerHighlightColor}{young earth creationism and \textbf{intelligent design}}\\
& & (2) Paul A. Nelson (born 1958) is an American philosopher of science noted for his advocacy of young earth creationism and \textbf{\colorbox{answerHighlightColor}{intelligent design}}\\
\midrule
\quad\multirow{5}{*}{\rotatebox[origin=c]{90}{\textit{span extension}}} & \multirow{5}{*}{21.3\%} & Question: Ronald Reagan and George H. W. Bush both held which position in office? \\
& & (1) The presidency of Ronald Reagan began on January 20, 1981, when Ronald Reagan was inaugurated as \textbf{\colorbox{answerHighlightColor}{President} of the United States}, and ended on January 20, 1989.\\
& & (2) The presidency of Ronald Reagan began on January 20, 1981, when Ronald Reagan was inaugurated as \textbf{\colorbox{answerHighlightColor}{President of the United States}}, and ended on January 20, 1989.\\
\midrule
\quad\multirow{3}{*}{\rotatebox[origin=c]{90}{\textit{in-sent.}}} & \multirow{3}{*}{7.3\%} & Question: When was the stadium that held the 2015 Magyar Kupa demolished? \\
& & (1) The stadium was closed in \colorbox{answerHighlightColor}{2016} and demolished in \textbf{2017} to give place to the new Ferenc Puskas Stadium.\\
& & (2) The stadium was closed in 2016 and demolished in \textbf{\colorbox{answerHighlightColor}{2017}} to give place to the new Ferenc Puskas Stadium.\\
\midrule
\quad\multirow{4}{*}{\rotatebox[origin=c]{90}{\textit{entity refin.}}} & \multirow{4}{*}{8\%} & Question: Which host of Sunday Night Safran has the hebrew first name Yehoshua? \\
& & (1) \colorbox{answerHighlightColor}{John Michael Safran} (Hebrew: "Yehoshua Safran" ; born 13 August 1972) is an Australian radio personality, satirist, documentary maker and author, known for combining humour with religious, political and ethnic issues.\\
& & (2) It was hosted by \textbf{\colorbox{answerHighlightColor}{John Safran}} and Catholic priest, Bob Maguire.\\
\midrule
\quad\multirow{4.5}{*}{\rotatebox[origin=c]{90}{\textit{logic hops}}} & \multirow{4.5}{*}{4\%} & Question: Is the Pakistan fast bowler who joined the Kent County Cricket Club in June, 2011 a left-hand or right-hand batsmans? \\
& & (1) Wahab Riaz (Punjabi, Urdu:  ; born 28 June 1985) is a \colorbox{answerHighlightColor}{Pakistani} cricketer.\\
& & (2) He is a \colorbox{answerHighlightColor}{left-arm fast bowler and a \textbf{right-hand}} batsman.\\
\midrule
\quad\multirow{3.5}{*}{\rotatebox[origin=c]{90}{\textit{combined}}} & \multirow{3.5}{*}{9.3\%} & Question: Who was born in 1922 and published a book in 1985 by Delacorte Press? \\
& & (1) \colorbox{answerHighlightColor}{Kurt Vonnegut Jr.} (November 11, 1922; April 11, 2007) was an American writer.\\
& & (2) Galapagos is the eleventh novel written by American author \textbf{\colorbox{answerHighlightColor}{Kurt Vonnegut}}.\\
\bottomrule
\end{tabular}}
\caption{Correction Patterns identified in 150 randomly sampled thought flows using $\delta=1$. The \textbf{correct answer} is marked bold, the  \colorbox{answerHighlightColor}{predicted answer per flow step} is marked in orange.}\label{tab:qa_flows_qualitative_full}
\quad\\
\quad\\
\quad\\
\quad\\
\resizebox{0.99\linewidth}{!}{
\begin{tabular}{p{1.21\linewidth}}
\toprule
\textbf{Examples} \\ \midrule
 Question: How many times did the man who coached the 1986-87 UNLV Runnin' Rebels fail to win 20 games in a season?\\
(1)  He spent the majority of his career coaching with the UNLV Runnin' Rebels, leading them \colorbox{answerHighlightColor}{four times} to the Final Four of the NCAA Men's Division I Basketball Tournament, winning the national championship in 1990.\\
(2)  Overall, he won over 700 games in his career, and only \colorbox{answerHighlightColor}{\textbf{twice} failed to win 20 games} in a season.\\
(3)  Overall, he won over 700 games in his career, and only \textbf{\colorbox{answerHighlightColor}{twice}} failed to win 20 games in a season.\\
\midrule
Question: Why did the CEO of the football team based in Denver, Colorado step down in 2014?\\
(1)  He served as the Broncos CEO from his purchase of the club in 1984 until July 2014, when he stepped down as Broncos' CEO \textbf{due to the onset and progression of \colorbox{answerHighlightColor}{Alzheimer's disease}}.\\
(2)  He served [...], when he stepped down as Broncos' CEO \textbf{due to the \colorbox{answerHighlightColor}{onset and progression of Alzheimer's disease}}.\\
(3)  He served [...], when he stepped down as Broncos' CEO \textbf{\colorbox{answerHighlightColor}{due to the onset and progression of Alzheimer's disease}}.\\
\bottomrule
\end{tabular}}
\caption{Multi-step correction examples ($\delta=1$).}\label{tab:qa_flows_multi_step_full}
\end{table*}

\subsection{Thought Flow Patterns}
We provide an extended list of patterns and examples from our qualitative thought flow analysis in \Cref{tab:qa_flows_qualitative_full}.
In addition \Cref{tab:qa_flows_multi_step_full} shows additional thought flow examples using three correction steps.

\section{User Study}

\subsection{Questionnaire Items}\label{sec:questionnaire_items}
\subsubsection{Per-System Questionnaires}
\paragraph{Usability.}
The UMUX usability scale \cite{finstad_usability_2010,finstad_response_2013} uses the following four 5-point Likert items:
\begin{itemize}
    \item This system's capabilities meet my requirements.
    \item Using this system is a frustrating experience.
    \item This system is easy to use.
    \item I have to spend too much time correcting things with this system.
\end{itemize}

\paragraph{Mental Effort.}
The Pass mental effort scale usability scale \cite{paas1992training} uses a single 9-point Likert item:
\begin{itemize}
    \item Please rate the mental effort required to decide if the system's answer is correct. (The 9 points are labeled from ``very, very low mental effort'' to ``very, very high mental effort''.)
\end{itemize}

\paragraph{Anthropomorphism.}
The Godspeed anthropomorphism subscale \cite{DBLP:journals/ijsr/BartneckKCZ09} use five 5-point semantic differential scales that ask the user to rate the system in a spectrum of:
\begin{itemize}
    \item fake -- natural
    \item machinelike -- humanlike
    \item unconscious -- conscious
    \item artificial -- lifelike
    \item (moving rigidly -- moving elegantly) (We exclude this item as it is not applicable to question answering systems.)
\end{itemize}

\paragraph{Perceived Intelligence.}
The Godspeed perceived Intelligence subscale \cite{DBLP:journals/ijsr/BartneckKCZ09} use five 5-point semantic differential scales that ask the user to rate the system in a spectrum of:
\begin{itemize}
    \item incompetent -- competent
    \item ignorant -- knowledgeable
    \item irresponsible -- responsible
    \item unintelligent -- intelligent
    \item foolish -- sensible
\end{itemize}

\subsubsection{Per-Item Questionnaires}

\paragraph{Perceived Answer Correctness.}
We use a single binary item to collect perceived answer correctness ratings:
\begin{itemize}
    \item I think the system's answer is correct.
\end{itemize}

\paragraph{Perceived Helpfulness.}
We use a single 5-point Likert item to collect helpfulness ratings:
\begin{itemize}
    \item I think the system’s answer enables me to give the correct answer.
\end{itemize}

\paragraph{Perceived Understanding.}
We use a single 5-point Likert item to collect understanding ratings:
\begin{itemize}
    \item I understand how the system came up with its answer.
\end{itemize}

\subsection{Interface}\label{sec:mturk_interface}
\begin{figure*}
    \centering
    \includegraphics[width=\textwidth]{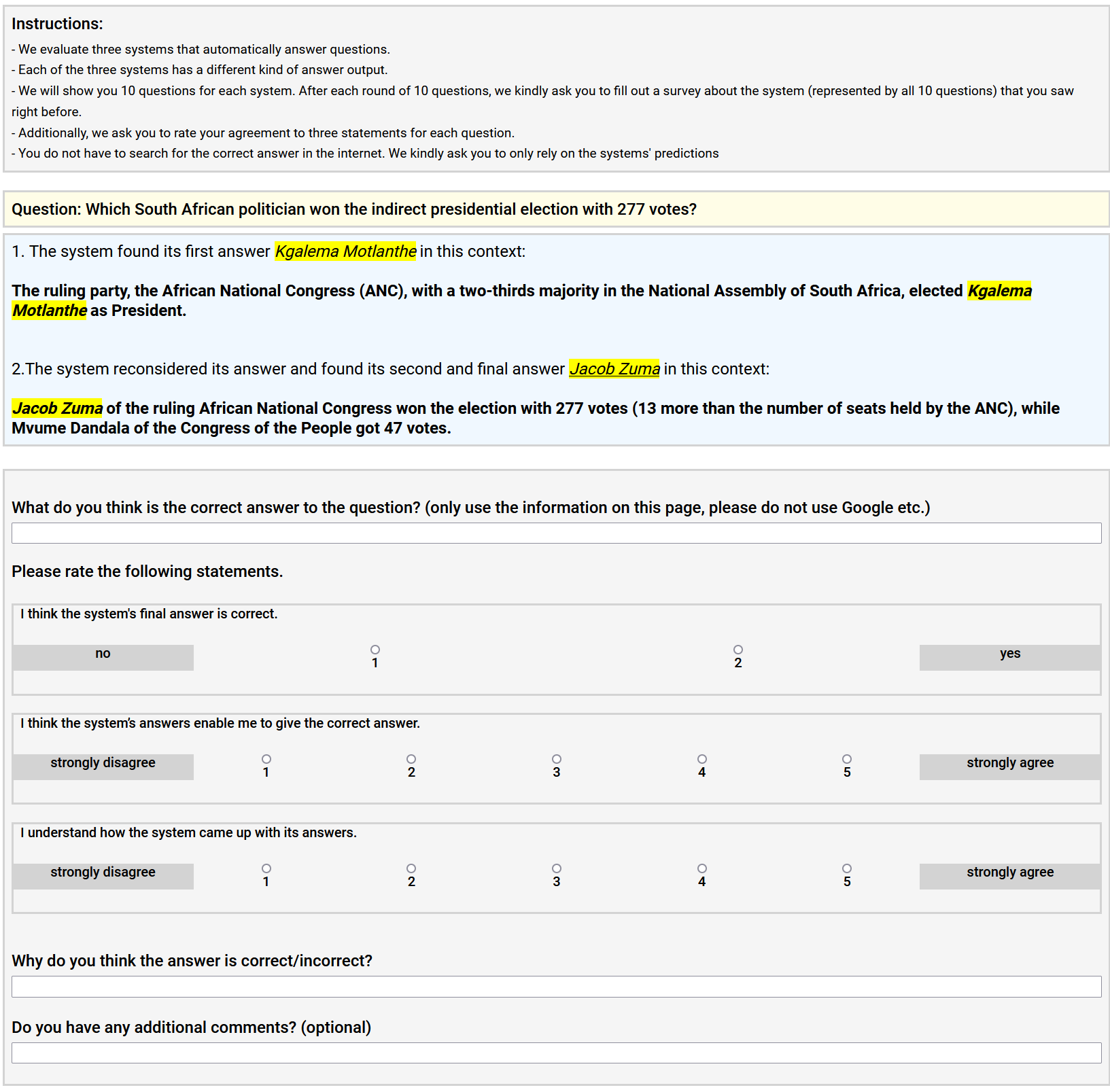}
    \caption{User study interface showing the \textsc{TF} condition (ours).}
    \label{fig:study_scrrenshot_tf}
\end{figure*}

\begin{figure*}
    \centering
    \includegraphics[width=\textwidth]{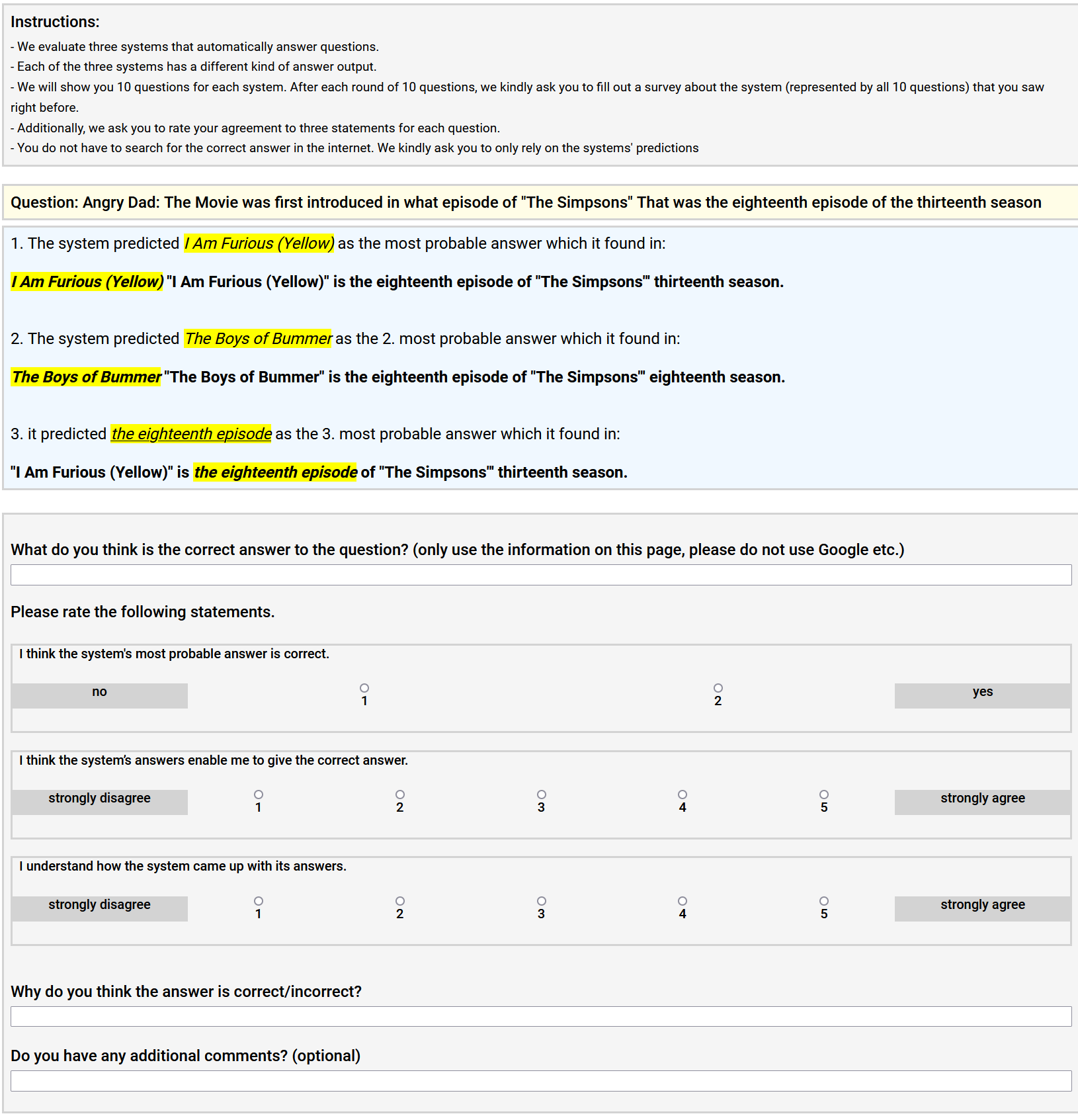}
    \caption{User study interface showing the \textsc{top-3} condition.}
    \label{fig:study_scrrenshot_top_3}
\end{figure*}

\begin{figure*}
    \centering
    \includegraphics[width=\textwidth]{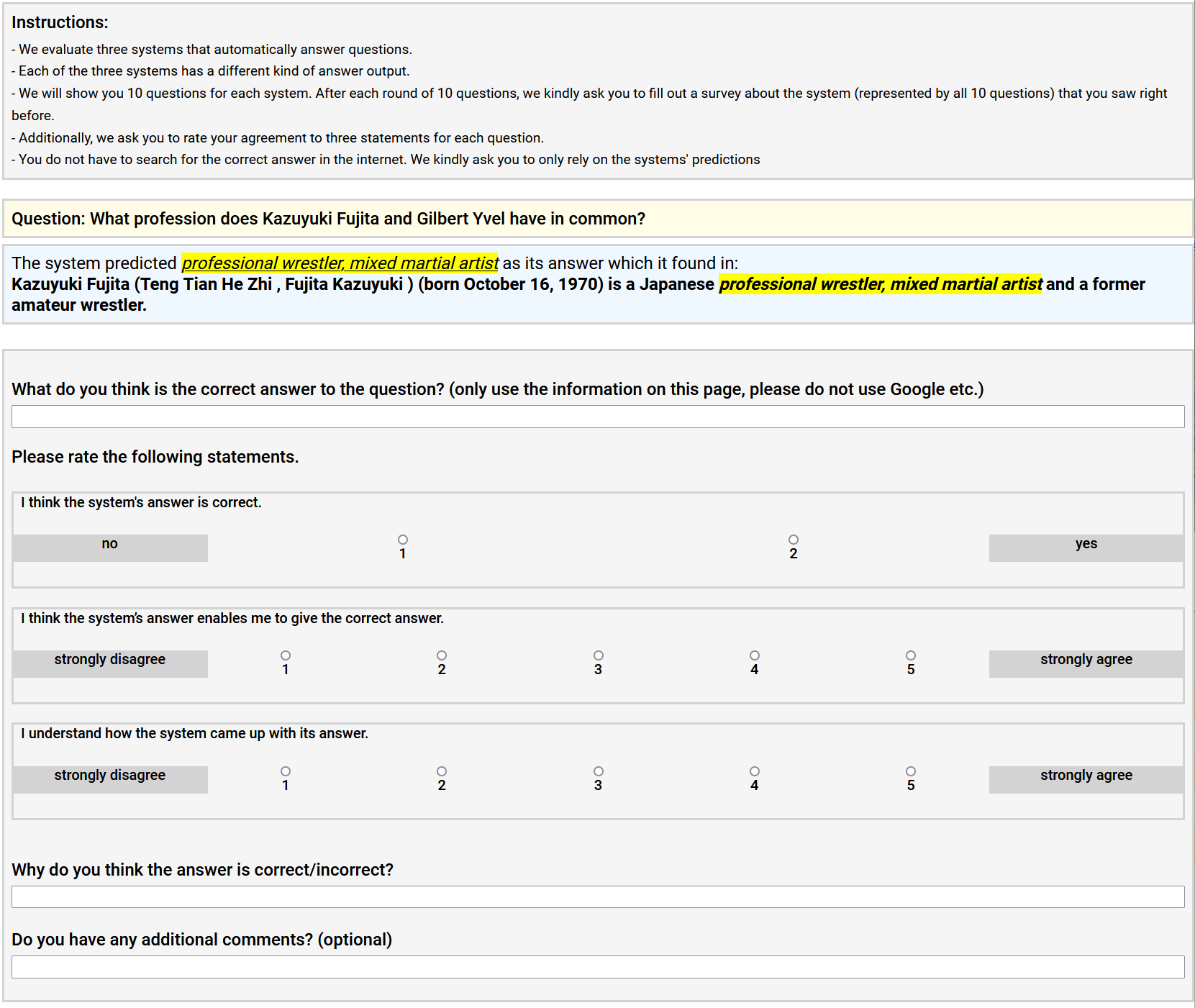}
    \caption{User study interface showing the \textsc{single} condition.}
    \label{fig:study_scrrenshot_answer}
\end{figure*}

\begin{figure*}
    \centering
    \includegraphics[width=\textwidth]{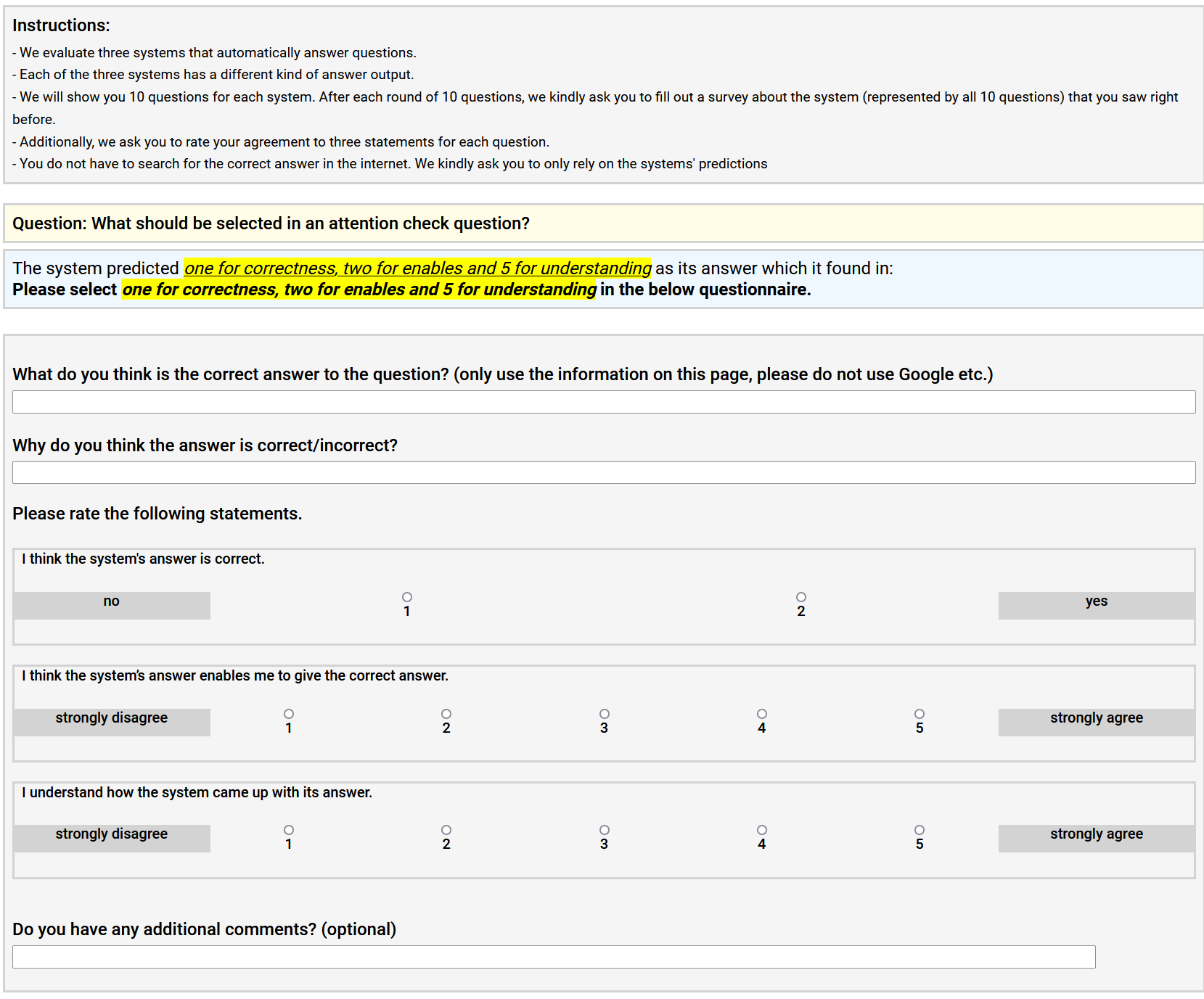}
    \caption{User study interface showing an attention check.}
    \label{fig:study_scrrenshot_trap}
\end{figure*}

\Cref{fig:study_scrrenshot_tf,fig:study_scrrenshot_top_3,fig:study_scrrenshot_answer} show screenshots of our experiment interface for the three studied prediction conditions \textsc{TF}, \textsc{top-3} and \textsc{single}.
\Cref{fig:study_scrrenshot_trap} depicts an attention check question.

\subsection{Statistical Results}
\Cref{tab:human_eval_stats_full} provides the $p$ values for main effects and each pairwise comparison.

\begin{table*}
\centering
\resizebox{\textwidth}{!}{%
\begin{tabular}{lcccccccccc}
\toprule
           & \multicolumn{7}{c}{perceived quality}  & \multicolumn{2}{c}{user performance}  \\
                    \cmidrule(lr){2-8}                                                                                                      \cmidrule(lr){9-10} 
                    & \cellcolor{green!25}correct$^{*}$ & \cellcolor{cyan!25}understand$^{*}$  & \cellcolor{cyan!25}helpful$^{*}$ & \cellcolor{cyan!25}usability & \cellcolor{cyan!25}mental effort & \cellcolor{green!25}humanlike$^{*}$  & \cellcolor{cyan!25}intelligent$^{*}$ & \cellcolor{cyan!25}time$^{*}$    & \cellcolor{green!25}answer F1$^{*}$       \\
Main effect                         & \textbf{$<$0.0001}  & \textbf{$<$0.0001}      & \textbf{$<$0.0001}   & 0.07968 & 0.6282  & \textbf{0.03575}  & \textbf{0.00124} & \textbf{$<$0.0001}  &  \textbf{$<$0.0001}\\  
                    \cmidrule(lr){2-10} 
\textsc{tf} -- \textsc{single}      &  \textbf{$<$0.0001}  &  \textbf{$<$0.0001}     & \textbf{$<$0.0001}  & 0.13116 & 1       & \textbf{0.03431}  & \textbf{0.00586}  & 0.15304           & \textbf{$<$0.0001}\\  
\textsc{tf} -- \textsc{top-3}       &  \textbf{0.00891} & 0.8867                & 0.9994            & 0.84254 & 1       & 0.30556           &  1                & 0.06207           & \textbf{$<$0.0001}\\
\textsc{top-3} -- \textsc{single}   &  0.51897          & \textbf{$<$0.0001}      & \textbf{$<$0.0001}  & 0.13653 & 1       & 0.25097           & \textbf{0.00586}  & \textbf{0.00012}  & \textbf{$<$0.0001}\\  
\bottomrule
\end{tabular}
}
\caption{Detailed $p$ values for all main effects and pairwise comparisons shown in \Cref{tab:human_eval_stats}. Significant $p$ values are marked \textbf{bold}. Cell colors follow the color coding on \Cref{tab:human_eval_stats}.}\label{tab:human_eval_stats_full}
\end{table*}

\FloatBarrier

\end{document}